\documentclass[letterpaper]{article} 
\usepackage{aaai23}  
\usepackage{times}  
\usepackage{helvet}  
\usepackage{courier}  
\usepackage[hyphens]{url}  
\usepackage{graphicx} 
\usepackage{natbib}  
\usepackage{caption} 
\usepackage{algorithm}
\usepackage{algorithmic}

\usepackage{amssymb}
\usepackage{amsmath}
\usepackage{booktabs}
\usepackage{enumitem}
\usepackage{multirow}
\usepackage{amsthm,amsmath,amssymb}
\usepackage{mathrsfs}

\usepackage{xcolor}

\newcommand{\model}{AnKGE} 

\usepackage{newfloat}
\usepackage{listings}
\DeclareCaptionStyle{ruled}{labelfont=normalfont,labelsep=colon,strut=off} 
\floatstyle{ruled}
\newfloat{listing}{tb}{lst}{}
\floatname{listing}{Listing}
\title{Analogical Inference Enhanced Knowledge Graph Embedding \\}
\author{
    Zhen Yao\textsuperscript{\rm 1}\equalcontrib, 
    Wen Zhang\textsuperscript{\rm 1}\equalcontrib, 
    Mingyang Chen\textsuperscript{\rm 2}, 
    Yufeng Huang\textsuperscript{\rm 1}, 
    Yi Yang\textsuperscript{\rm 4}, 
    Huajun Chen\textsuperscript{\rm 2,3,5}\thanks{Corresponding Author.}
}
\affiliations{
    \textsuperscript{\rm 1}School of Software Technology, Zhejiang University\\
    \textsuperscript{\rm 2}College of Computer Science and Technology, Zhejiang University\\
    \textsuperscript{\rm 3}Donghai Laboratory, Zhoushan 316021, China\\ 
    \textsuperscript{\rm 4}Huawei Technologies Co., Ltd\\
    \textsuperscript{\rm 5}Alibaba-Zhejiang University Joint Institute of Frontier Technologies\\

    \{yz0204, zhang.wen, mingyangchen, huangyufeng, huajunsir\}@zju.edu.cn \\
    yangyi193@huawei.com, 
}

\usepackage{bibentry}

\begin{document}

\maketitle

\begin{abstract}
Knowledge graph embedding (KGE), which maps entities and relations in a knowledge graph into continuous vector spaces, has achieved great success in predicting missing links in knowledge graphs. However, knowledge graphs often contain incomplete triples that are difficult to inductively infer by KGEs. To address this challenge, we resort to analogical inference and propose a novel and general self-supervised framework AnKGE to enhance KGE models with analogical inference capability. We propose an analogical object retriever that retrieves appropriate analogical objects from entity-level, relation-level, and triple-level. And in AnKGE, we train an analogy function for each level of analogical inference with the original element embedding from a well-trained KGE model as input, which outputs the analogical object embedding. In order to combine inductive inference capability from the original KGE model and analogical inference capability enhanced by AnKGE, we interpolate the analogy score with the base model score and introduce the adaptive weights in the score function for prediction. Through extensive experiments on FB15k-237 and WN18RR datasets, we show that AnKGE achieves competitive results on link prediction task and well performs analogical inference. 
\end{abstract}

\section{Introduction}

Knowledge graphs (KGs) storing a large number of triples in the form of \textit{(head entity, relation, tail entity)}, $(h,r,t)$ for short, are popular data structures for representing factual knowledge.
Many knowledge graph projects such as Freebase \cite{DBLP:freebase}, WordNet \cite{DBLP:wordnet}, YAGO \cite{DBLP:YAGO} and DBpedia \cite{DBLP:DBpedia} are significant foundations to support artificial intelligence applications. 
They have been successfully used in downstream applications such as word sense disambiguation \cite{DBLP:WordSD}, question answering \cite{DBLP:QA}, and information extraction \cite{DBLP:Information}, gaining widespread attention. However, most KGs are incomplete, so predicting the missing links between entities is a fundamental problem for KGs called link prediction. 
One of the common approaches to this problem is knowledge graph embedding (KGE) methods, which make prediction through a predefined triple score function with learnt entity and relation embeddings as input.
Many KGE models have been proposed like TransE \cite{DBLP:TransE}, DistMult \cite{DBLP:DistMult}, RotatE \cite{DBLP:RotatE} and HAKE \cite{DBLP:HAKE}. These methods have gained great success in knowledge graph completion task.

For most KGE methods, the parametric learning paradigm can be viewed as \emph{memorization} regarding training data as a book and predicting missing links as the close-book test \cite{DBLP:REkNN}, which belongs to inductive inference. 
However, the large knowledge graphs often contain incomplete triples that are difficult to be inductively inferred by applying \emph{memorization} paradigm. 
Nevertheless, the problem may be well solved by using analogical inference method. That is because analogical inference is a referential method, which retrieves similar solutions to solve new problems, similar to an open-book examination.
For example, it seems that most people could not remember even learn about what company Ron Wayne founded. However, if they know that Ron Wayne and Steve Jobs are the co-founders, i.e., Steve Jobs and Ron Wayne are analogical objects in this context, and it is well known that Steve Jobs founded Apple Inc.; thus they could analogically infer that Ron Wayne founded Apple Inc. .

In order to enhance KGEs with analogical inference capability, 
there are three problems should be solved: 
1) How to define the analogical objects of elements given a task? 
2) How to enable the model to map elements to analogical objects? 
3) How to combine the original inductive inference capability and enhanced analogical inference capability? 

We propose AnKGE, a novel and general self-supervised framework, which solves these problems very well and enhances well-trained KGEs with analogical inference capability.
For problem 1, we think that an analogical object can solve the given task well, and inspired by the nearest neighbor language model \cite{DBLP:kNN-LM}, we propose an analogical retriever covering objects of three levels, including entity, relation, and triple level. Specifically, we consider the score function of KGEs as the assessment of the quality of triples and regrade the replacement triples with the highest scoring as the appropriate analogical objects.
For problem 2, we trained a projecting function using analogical objects as supervision signals. This function projects original objects onto appropriate analogical objects.
For problem 3, we interpolate the analogy score with the base model score to combine the original inductive inference capability and enhanced analogical inference capability. Moreover, we introduce the adaptive weight to adjust analogical inference in knowledge graph completion task.

Finally, through link prediction experiments on FB15k-237 and WN18RR datasets, we demonstrate the AnKGE is significantly compatible and outperforms the other baseline models.
To the best of our knowledge, AnKGE is the first framework to enhance KGEs with analogical inference ability.

In summary, our contributions in this work include:
\begin{itemize}
\item We explore the knowledge graph completion task from the analogical inference view. We propose an effective retrieval method covering three levels to obtain the appropriate analogy objects.

\item We propose a novelty analogical inference enhanced framework called AnKGE, which could project original objects onto appropriate objects for analogical inference. To our knowledge, the AnKGE is the first framework of knowledge graph embedding to enhance analogical inference ability.

\item We conduct experimental evaluations to demonstrate the proposed AnKGE is significantly compatible and achieves competitive performance on FB15k-237 and WN18RR datasets, promising practical applications.
\end{itemize}

\section{Related Work}
\subsubsection{Knowledge graph embedding}
\label{sec:kge model cate}
According to previous work \cite{DBLP:NeuralKG}, the KGE methods can be divided into two categories based on the scoring function and whether a global graph structure is utilized. The first category is the Conventional KGEs (C-KGEs), which apply a geometric assumption in vector space for true triples and use single triple as input for triple scoring. Conventional KGEs use the score function to measure the plausibility of triple. TransE \cite{DBLP:TransE} is a representative conventional KGE method whose score function is $ \left\| \mathbf{h+r-t} \right\|_2 $. What is more, there are many variants to improve the performance of TransE, such as RotatE \cite{DBLP:RotatE}, DistMult \cite{DBLP:DistMult} and HAKE \cite{DBLP:HAKE}. The other category is the GNN-based methods, which use representations of entities and relations aggregated from their neighbors in the graph instead of embedding them for triple scoring to capture the graph patterns explicitly. R-GCN \cite{DBLP:R-GCN} is the first GNN framework to model relational data. It introduces relation-specific transformations when neighbor aggregating. 
SE-GNN \cite{DBLP:SE-GNN} models three levels semantic evidence into knowledge embedding. Note that SE-GNN introducing three levels from the semantic evidence view differs from our three levels analogical objects.

\subsubsection{Enhanced KGE framework}
Recently, some work has proposed some frameworks and strategies to improve the performance of KGE models, which are called enhanced KGE, such as CAKE \cite{DBLP:CAKE}, PUDA\cite{DBLP:PUDA} and REP \cite{DBLP:REP}. CAKE is a commonsense-aware knowledge embedding framework to extract commonsense from factual triples with entity concepts automatically, which generates commonsense augments to facilitate high-quality negative sampling.
PUDA is a data augmentation strategy to address the false negative
and data sparsity issue. 
REP is a post-processing technique to adapt pre-trained KG embeddings with graph context. Our method is designed to enhance a well-trained KGE model with analogical inference capability belonging to the enhanced KGE framework.

\subsubsection{Analogical inference}
In classic artificial intelligence, analogical inference was an active research topic. However, the early computational model of analogy-making study \cite{DBLP:analogytheore, DBLP:analogytheore2} mainly focuses on structure mapping theory and its implementation in the structure mapping engine.
Recently, some researchers proposed k-Nearest Neighbor language model(kNN-LM) \cite{DBLP:kNN-LM}, 
which can directly query training examples at test time, also can be considered the analogy inference model in the neural language process topic. 
While effective, these models often require retrieval from a large datastore at test time, significantly increasing the inference overhead.
In the field of knowledge graph, the study of analogical inference to solve knowledge graph incomplete problem is missing.
ANALOGY \cite{DBLP:Analogy} is the first method for modeling analogical structures in multi-relational embedding, but the performance is not good.
Differ in our method uses the nearest neighbor method to perform explicit analogy, 
ANALOGY uses the commutativity constraint of the normal matrix to model analogical relations implicitly.

\section{Analogical Object Retriever}
\label{sec: Retriever}
Before introducing our method, in this section, we firstly introduce the background of knowledge graph and analogical inference, and then we propose the analogical object retrievers that retrieve appropriate analogical objects from entity-level, relation-level, and triple-level. The retrieved analogical objects will be used as supervision signals with our method. 

\paragraph{Background}
A knowledge graph is denoted as $\mathcal{G}=(\mathcal{E}, \mathcal{R}, \mathcal{F})$, where $\mathcal{E}$ represents the set of entities, $\mathcal{R}$ represents the set of relations, and $\mathcal{F}=\{(h, r, t)\} \subseteq \mathcal{E} \times \mathcal{R} \times \mathcal{E}$ represents the set of triple facts. 

Analogical inference, which has been long researched in artificial intelligence, maps the target problem to a known source problem that could effectively utilize known knowledge \cite{DBLP:journals/ai/Hall89}.
Applying analogical inference into link prediction task $(h,r,?)$ in knowledge graphs, instead of directly predicting the tail entity $t$, we could make prediction through similar triples that we know, i.e. triples in train dataset. 
We consider similar triples are composed by analogical objects of $(h,r,t)$. Specifically, we assume that the analogy objects may come from three levels: the analogy of head entity $h$ part resulting similar triple $(h', r, t)$(entity-level), the analogy of relation $r$ part resulting similar triple $(h, r', t)$ (relation-level) and the analogy of combination pair $(h, r)$ part $t$ resulting similar triple $(h', r', t)$ (triple-level). 

Thus, we propose three retrievers to obtain different level's analogical objects.

\paragraph{Entity-Level Retriever}
The retriever is designed based on the score function $f_{kge}(h,r,t)$ predefined in a well-trained KGE model, where triples with higher scores are assumed with higher probability to be true.
Inspired by the nearest neighbor language model \cite{DBLP:kNN-LM}, we replace all possible objects of the triple and regrade the replacement triples with highest scoring as the appropriate analogical objects.
Given a triple $(h,r,t)$, entity-level retriever retrieves similar true triples $(h',r,t)$ for entity-level analogical inference. For example, we could get the answer of $(Sergey~Brin, found, ?)$ is $Google$ through $(Larry~Page, found, Google)$ if we know Sergey Brin and Larry Page are co-founders. 

Specifically, in entity-level retriever, we first replace $h$ with all entities resulting $|\mathcal{E}|$ replacement triples, and then regard triples with highest scores measured by the KGE as similar triples. And we name the head entity in similar triples as analogical objects from entity-level retriever. Thus analogical object set could be represented as
\begin{equation}
    \mathsf{E}_{N_e}^{hrt}=\{{h}_{i} ~|~ Top(~\{f_{kge}( \mathbf{{h_{i},r,t}} )~|~ h_{i} \in \mathcal{E}\}~)_{N_{e}}\},
\end{equation}
where  $Top( \cdot )_{k}$ denotes the $k$ elements with top $k$ values among all inputs, ${f}_{kge}(\cdot,\cdot,\cdot)$ is the predefined score function in KGE model, and \emph{hrt} denotes a specific triple $(h,r,t)$ as input. If not otherwise specified, we omit \emph{hrt} and use ${E}_{N_e}$ instead of ${E}_{N_e}^{hrt}$ for simplicity.
Compared to retrieving similar triples directly from the train dataset, retrieving according to scores from KGEs could help overcome the incompleteness of KGs.

\paragraph{Relation-Level Retriever}
Given $(h,r,t)$, relation-level retriever retrieves $(h,r',t)$ for relation-level analogical inference, since there are relations with similar contexts in KGs. For example, the founder of a company is usually the board member. Thus the relation-level analogy object of $found$ is $board\_member$.
Similar to the entity-level retriever, the analogical object set of $(h,r,t)$ from relation-level retriever is as follow :
\begin{equation}
    \mathsf{R}_{N_r}=\{{r}_{i} ~|~ Top(~\{f_{kge}( \mathbf{h,r_{i},t} )~|~ r_{i} \in \mathcal{R}\}~)_{N_{r}}\}.
 \label{equ: relation level retrieve}
\end{equation}

\paragraph{Triple-Level Retriever}
Given $(h,r,t)$, triple-level retriever retrieves $(h', r', t)$ for triple-level analogical inference, which is the combination of entity-level and relation-level retriever. 
For instance, Sergey Brin is the founder of Google and Sundar Pichai is the CEO of Google. Therefore, the triple-level analogical objects of $(Sergey Brin, found)$ is $(Sundar Pichai, CEO)$. 
Actually, the number of all candidate $(h',r')$ pairs is in millions in most knowledge graphs.
In order to reduce the cost of retrieving candidate pairs and inspired by the principle of locality, we often select \emph{m} entities and \emph{n} relations with high triple scores separately, and then pair them with each other. 
Thus the set of analogical objects, namely $(h', r')$ pairs, from triple-level retriever is
\begin{equation}
\begin{aligned}
    \mathsf{T}_{N_t} &= \{({h}_{i},{r}_{i})~| \\ &~ Top(~\{f_{kge}( \mathbf{h_{i},r_{i},t} )~|~ h_{i}\in \mathsf{E}_m, r_{i}\in \mathsf{R}_n\})_{N_{t}}\}.
\label{equ: triple level retrieve}
\end{aligned}
\end{equation}

\section{Methodology}
\begin{figure*}[t]
    \centering
\includegraphics[scale=0.5]{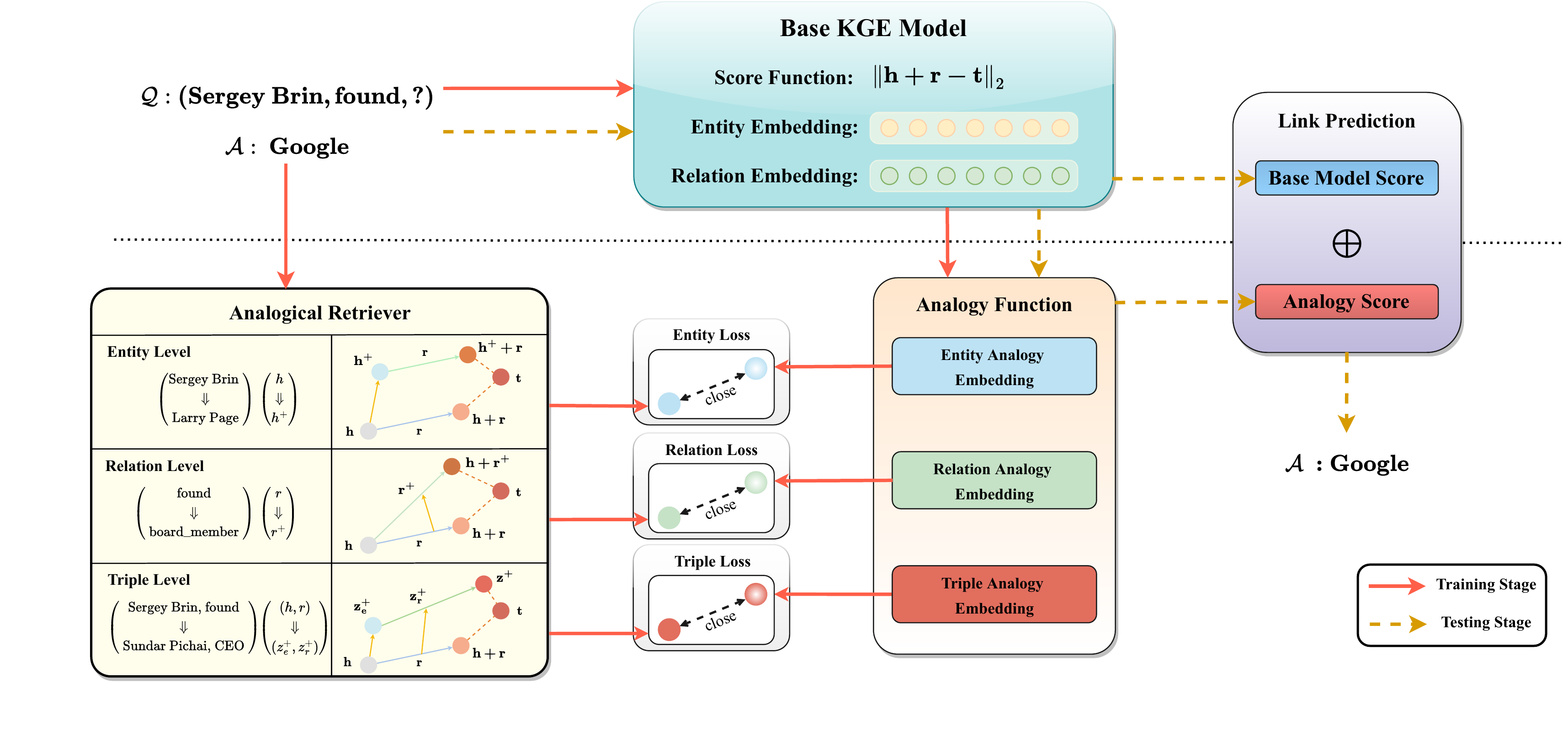}
	\caption{This is the AnKGE structure diagram with TransE as the base model. For simplicity, we set the numbers of three levels analogical object are 1. The upper half of figure shows the module of base model. The predefined score function is applied to learnt embedding to get the well-trained model. The lower half of figure shows the module of AnKGE. First, AnKGE retrieves the analogy objects for training the analogy function. The solid line arrow indicates the AnKGE training process. Then, AnKGE remakes the prediction ranking by interpolating analogy score. The dashed line arrow indicates the AnKGE testing process.}
	
	\label{fig:overview}
\end{figure*}

In this section, we present a novel KGE enhanced framework called Analogy Enhanced Knowledge Graph Embedding (\model), which could model the three levels of analogical inference as introduced in Section \ref{sec: Retriever}.
Next, we first introduce the definition of analogy function (Section \ref{sec: ana emb}) and how to train it by using analogical objects (Section \ref{sec: agg emb} and Section \ref{sec:loss}). Finally, we introduce how to combine the original inductive inference capability and enhanced analogical inference capability in knowledge graph completion task. (Section \ref{sec:LP})

\subsection{Analogy Function}
\label{sec: ana emb}

Given a well-trained KGE model $\mathcal{M}= \{ \mathbf{E}, \mathbf{R}, f_{kge}, \Theta \} $, where $\mathbf{E}, \mathbf{R}$ and $f_{kge}$ are entity embedding table, relation embedding table, and score function of the $\mathcal{M}$, and $\Theta$ is the set of other parameters,  \model\;  enhances $\mathcal{M}$ with capability of analogical inference through a projecting function called \textit{analogy function $f$}.
We train an analogy function for each level of analogical inference with the original element embedding from $\mathbf{E}$ or $\mathbf{R}$ in $\mathcal{M}$ as input and output the analogical object embedding to conduct link prediction.

Specifically, analogy function for relation-level analogical inference $f_{rel}$ maps an original embedding of a relation $r$ in $(h,r,t)$ to the analogical embedding through a relation projecting vector $\mathbf{v}^{R}_{r} \in \mathbb{R}^{d_r}$ that
\begin{equation}
    f_{rel}(r) = \mathbf{r_{a}} = 
    \mathbf{v}^{R}_{r}
    \circ \mathbf{r},
\end{equation}
where $d_r$ is the relation hidden dimension, $\circ$ is the element-wise product.

Similarly, the analogy function for entity-level analogical inference $f_{ent}$ maps an original embedding of an entity $h$ in $(h,r,t)$ to the analogical embedding. Considering that an entity generally tends to be associated with multiple relations, we define $f_{ent}$ as:
\begin{equation}
    f_{ent}(h, r) = \mathbf{h_{a}} = \mathbf{v}^{E}_{h} \circ \mathbf{h} + \lambda \mathbf{M}_{trans} \times  \mathbf{v}^{R}_{r} \circ \mathbf{r},
    \label{equ: ent ana fun}
\end{equation}
where 
$\mathbf{v}^{E}_{h} \in \mathbb{R}^{d_e}$ is the entity projecting vector and 
$d_e$ is the entity hidden dimension. $\mathbf{M}_{trans} \in \mathbb{R}^{d_e \times d_r}$ denotes the transformation matrix that enable to make relation $r$ into consideration. $\lambda$ is a weight hyper-parameter.

Analogy function for triple-level analogical inference $f_{trp}$ outputs the analogical embedding of entity and relation pairs through combining embedding of entity-level and relation-level according to KGEs as follows:
\begin{equation}
 f_{trp}(h,r) =  \mathbf{z_{a}} = g_{kge}\left( \mathbf{h_{a}, r_{a}} \right),
\end{equation}
${g}_{kge}(\cdot,\cdot)$ is the function in KGEs that maps a head entity embedding to the tail entity embedding according to the given relation embedding.
$g_{kge}(\cdot,\cdot)$ and $f_{kge}(\cdot, \cdot, \cdot)$ of representative KGE models are provided in Appendix \ref{app:score function}.
\begin{table*}[t]
\centering
\small
\begin{tabular}{lllll|llll}
\toprule
& \multicolumn{4}{c}{\textbf{FB15k-237}} & \multicolumn{4}{c}{\textbf{WN18RR}} \\ 
\cmidrule(lr){2-5} \cmidrule(lr){6-9}
& \multicolumn{1}{c}{\textbf{MRR}} & \multicolumn{1}{c}{\textbf{Hit@1}} & \multicolumn{1}{c}{\textbf{Hit@3}} & \multicolumn{1}{c}{\textbf{Hit@10}} & \multicolumn{1}{c}{\textbf{MRR}} & \multicolumn{1}{c}{\textbf{Hit@1}} & \multicolumn{1}{c}{\textbf{Hit@3}} & \multicolumn{1}{c}{\textbf{Hit@10}} \\ \midrule
\textbf{Conventional KGE} \\
TransE \cite{DBLP:TransE}
& 0.317 & 0.223 & 0.352 & 0.504 
& 0.224 & 0.022 & 0.390 & 0.520 \\
ANALOGY \cite{DBLP:Analogy}
& 0.256 & 0.165 & 0.290 & 0.436
& 0.405 & 0.363 & 0.429 & 0.474 \\
RotatE \cite{DBLP:RotatE}
& 0.336 & 0.244 & 0.370 & 0.524 
& 0.473 & 0.428 & 0.491 & 0.564 \\
HAKE \cite{DBLP:HAKE}
& 0.349 & 0.252 & 0.385 & 0.545 
& \underline{0.496} & \underline{0.452} & \underline{0.513} & 0.580 \\
Rot-Pro \cite{DBLP:Rot-Pro}
& 0.344 & 0.246 & 0.383 & 0.540
& 0.457 & 0.397 & 0.482 & 0.577 \\
PairRE \cite{DBLP:PairRE}
& 0.348 & 0.254 & 0.384 & 0.539 
& 0.455 & 0.413 & 0.469 & 0.539 \\
DualE \cite{DBLP:Dual}
& 0.365 & 0.268 & 0.400 & 0.559 
& 0.492 & 0.444 & \underline{0.513} & 0.584 \\ 
\midrule
\textbf{GNN-based KGE} \\
R-GCN \cite{DBLP:R-GCN}
& 0.249 & 0.151 & 0.264 & 0.417
& -     & -     & -     & -     \\
A2N \cite{DBLP:A2N}
& 0.317 & 0.232 & 0.348 & 0.486 
& 0.450 & 0.420 & 0.460 & 0.510 \\
CompGCN \cite{DBLP:CompGCN}
& 0.355 & 0.264 & 0.390 & 0.535
& 0.479 & 0.443 & 0.494 & 0.546 \\
SE-GNN \cite{DBLP:SE-GNN}
& 0.365 & \underline{0.271} & 0.399 & 0.549 
& 0.484 & 0.446 & 0.509 & 0.572 \\
\midrule
\textbf{Enhanced KGE} \\
CAKE \cite{DBLP:CAKE}
& 0.321 & 0.226 & 0.355 & 0.515
& -     & -     & -     & -     \\
PUDA \cite{DBLP:PUDA}
& \underline{0.369} & 0.268 & \underline{0.408} & \textbf{0.578}
& 0.481 & 0.436 & 0.498 & 0.582 \\
REP \cite{DBLP:REP}
& 0.354 & 0.262 & 0.388 & 0.540 
& 0.488 & 0.439 & 0.505 & \textbf{0.588} \\
\midrule

AnKGE-HAKE(ours)
& \textbf{0.385} & \textbf{0.288} & \textbf{0.428} & \underline{0.572}
& \textbf{0.500} & \textbf{0.454} & \textbf{0.515} & \underline{0.587} \\ 

\bottomrule
\end{tabular}
\caption{Link Prediction results on FB15k-237 and WN18RR. The best results are \textbf{bold} and second best results are \underline{underline}. }
\label{tab:link-prediction-sota}
\end{table*}

\subsection{Analogy Objects Aggregator}
\label{sec: agg emb}

In order to enhance the framework's robustness for analogical inference, we make the analogical objects retrieved following Section \ref{sec: Retriever} as the supervision signals for analogy functions. 
Specifically, we make the analogy embedding as introduced in Section \ref{sec: ana emb} to approach the weighted average of the analogical objects from KGE model $\mathcal{M}$.

The aggregated embeddings of entity-level and relation-level, $h^+$ and $r^+$ respectively, are calculated as follows
\begin{equation}
  \mathbf{h^+} = { \sum_{h_i \in \mathsf{E}_{Ne}}
    { \mathbf{h_i} ~ \mathcal{S} ({f_{kge}( \mathbf{ h_i,r,t)} }) }} ,
\end{equation}
\begin{equation}
  \mathbf{r^+} = {\sum_{r_i \in \mathsf{R}_{Nr}} 
    { \mathbf{r_i} ~ \mathcal{S} ({f_{kge}( \mathbf{h,r_i, t)} }) }} ,
\end{equation}
where $\mathcal{S}(\cdot)$ is the softmax function that converts a vector of $K$ real numbers into a probability distribution of $K$ possible outcomes, which is formulated as $\mathcal{S}\left(c_{i}\right)={e^{c_{i}}} / {\sum_{k=1}^{K} e^{c_{k}}}$.

Triple-level aggregated embedding $z^+$ is obtained by the firstly aggregating entity and relation embedding separately and then calculating combination embedding, which can be formulated as:
\begin{equation}
    \begin{split}
  \mathbf{z^+} =  & g_{kge}\left( \mathbf{z_e^+, z_r^+} \right),
    \\
  \mathbf{z_e^+} = & \sum_{ (h_i, r_i) \in \mathsf{T}_{N_t}} \mathbf{h_i} ~ \mathcal{S} ({f_{kge}( \mathbf{ {h_i,r_i,t}) } }),\\
  \mathbf{z_r^+} = & \sum_{ (h_i, r_i) \in \mathsf{T}_{N_t}} \mathbf{r_i} ~ \mathcal{S} ({f_{kge}( \mathbf{ {h_i,r_i,t}) } }).
    \end{split} 
\end{equation}

\subsection{Loss Function}
\label{sec:loss}
The training goal of the analogy function is to reduce the distance between the analogy embedding and aggregated embedding obtained following Section \ref{sec: ana emb} and \ref{sec: agg emb} respectively. 
In addition, considering that 
$f_{kge}$ performs priori on the 
truth value of triples, we take the analogy triple score as another supervision signal.
Therefore, given a pair of analogy embedding $\mathcal{X}_{a}$ and aggregated embedding $\mathcal{X}^{+}$ of a triple embeddings $(\mathbf{h,r,t})$, the loss function is
\begin{equation}
    \begin{split}
 \mathcal{L}(\mathcal{X}, & (\mathbf{h,r,t})) = \\
   & log\sigma\left( { \gamma\left\| { \mathcal{X}_{a} - \mathcal{X}^+ } \right\|_2 - f_{kge}( \mathbf{\mathbf{h,r,t}} )} \right) ,
 \label{equ: level loss}
    \end{split}
\end{equation}
where $\gamma$ is a hyper-parameter of the loss function, $\sigma$ is the sigmoid function. $\left\|\cdot\right\|_2$ is the euclidean norm. 

However, 
the three levels of analogical inference are not equally important for different triples. We add weight parameters for each loss of three levels and the final training objective is\footnote{During the gradient update, the parameters of the original model are frozen.}:
\begin{equation}
    \begin{split}
    \text{min}\;\; Loss = \sum_{(h,r,t)  \in  \mathcal{F}} \big( & \beta_{E}  \mathcal{L}(\mathbf{h}, \mathbf{(h_a,r,t)})  \\
     + & \beta_{R} ~ \mathcal{L}(\mathbf{r}, \mathbf{(h,r_a,t)} ) \\
     + & \beta_{T} ~ \mathcal{L}(\mathbf{z}, \mathbf{(h_a,r_a,t)}) \big) .
 \label{loss function}
    \end{split}
\end{equation}
As a result, considering the different contributions of three level, we introduce $\beta_E$, $\beta_R$ and $\beta_T$ to adjust gradient descent.
The three levels loss function distribution is positively correlated with the score of the analogy triple. Due to page limitation, we put the calculation details in Appendix \ref{app:loss}.

\subsection{Link Prediction}
\label{sec:LP}

For a test triple $(h,r,t)$ in test set $\mathcal{F}_{t e}$, we follow the kNN-LM \cite{DBLP:kNN-LM} and interpolate the analogy score with base model score to get the final score function:
\begin{equation}
\begin{aligned}
 Score(h,r,t) &=  f_{kge}\left( \mathbf{h,r,t} \right) +  \lambda_{E}f_{kge}\left(                    \mathbf{h_{a},r,t} \right) + 
      \\ & \lambda_{R}f_{kge}\left( \mathbf{h,r_{a},t} \right) + \lambda_{T}f_{kge}\left( \mathbf{h_{a},r_{a},t} \right)
 \label{score function}
\end{aligned}
\end{equation}
where $\lambda$ is the adaptive weight parameter, which is dynamically adjusts analogy weight according to training triples. 
$\lambda_{E}$ is proportional to the number of triples with the same $(r,t)$ in the training set. 
$\lambda_{R}$ is proportional to the number of triples with the same $(h,t)$ in the training set. 
$\lambda_{T}$ is proportional to the number of triples with the same tail entity in the training set. The formula for adaptive weight parameter is\footnote{When link prediction, we add reverse relations to expand the dataset and predict tail entity only, which is equivalent to the effect of predicting both head and tail entities. Each prediction will use all entities to replace tail entity. Thus, there is no risk of label leakage.}:
\begin{equation}
     \begin{split}
        &\lambda_{E} = {\min\left(  |\left\{ (h_{i},r,t) \in \mathcal{F}\right\}| / {N_{e}}, 1 \right)} \times \alpha_{E} ,
        \\
        &\lambda_{R} = {\min\left(  |\left\{ (h,r_{i},t) \in \mathcal{F}\right\}| / {N_{r}}, 1 \right)} \times \alpha_{R} ,
        \\
        &\lambda_{T} = {\min\left(|\left\{ (h_{i},r_{i},t) \in \mathcal{F}\right\}| / {N_{t}}, 1 \right)} \times \alpha_{T} ,
     \end{split}
 \label{adaptive weight}
\end{equation}
where ${\alpha_{E},\alpha_{R},\alpha_{T}}$ are basic weight hyper-parameters. Adaptive weight utilizes the train dataset to determine whether test triples are suitable for different levels of analogical inference. 
When all levels of analogical inference are not suitable, this score function degenerates to the base KGE model. In fact, AnKGE remakes the rank of hard-predicted triples in the base model by analogical inference to improve the prediction performance.

\begin{table*}[ht]
\centering
\small

\begin{tabular}{ccccc|cccc}
\toprule
& \multicolumn{4}{c}{\textbf{FB15k-237}} & \multicolumn{4}{c}{\textbf{WN18RR}} \\ 
\cmidrule(lr){2-5} \cmidrule(lr){6-9}
& \multicolumn{1}{c}{\textbf{MRR}} & \multicolumn{1}{c}{\textbf{Hit@1}} & \multicolumn{1}{c}{\textbf{Hit@3}} & \multicolumn{1}{c}{\textbf{Hit@10}} & \multicolumn{1}{c}{\textbf{MRR}} & \multicolumn{1}{c}{\textbf{Hit@1}} & \multicolumn{1}{c}{\textbf{Hit@3}} & \multicolumn{1}{c}{\textbf{Hit@10}} \\ \midrule

TransE
& 0.317 & 0.223 & 0.352 & 0.504 
& 0.224 & 0.022 & 0.390 & 0.520 \\
AnKGE-TransE
& \textbf{0.340} & \textbf{0.245} & \textbf{0.379} & \textbf{0.523}
& \textbf{0.232} & \textbf{0.031} & \textbf{0.402} & \textbf{0.526} \\

\midrule
RotatE
& 0.336 & 0.244 & 0.370 & 0.524 
& 0.473 & 0.428 & 0.491 & 0.564 \\
AnKGE-RotatE
& \textbf{0.366} & \textbf{0.273} & \textbf{0.405} & \textbf{0.546}
& \textbf{0.480} & \textbf{0.431} & \textbf{0.499} & \textbf{0.578} \\

\midrule
HAKE
& 0.349 & 0.252 & 0.385 & 0.545 
& 0.496 & 0.452 & 0.513 & 0.580 \\
AnKGE-HAKE
& \textbf{0.385} & \textbf{0.288} & \textbf{0.428} & \textbf{0.572}
& \textbf{0.500} & \textbf{0.454} & \textbf{0.515} & \textbf{0.587} \\

\midrule
PairRE
& 0.348 & 0.254 & 0.384 & 0.539 
& 0.455 & 0.413 & 0.469 & 0.539 \\
AnKGE-PairRE
& \textbf{0.376} & \textbf{0.281} & \textbf{0.417} & \textbf{0.558} 
& \textbf{0.462} & \textbf{0.415} & \textbf{0.480} & \textbf{0.556} \\

\bottomrule
\end{tabular}
\caption{AnKGE upon different model on FB15k-237 and WN18RR. The better results are \textbf{bold}.}
\label{tab:ankge-mult-model}
\end{table*}

\section{Experiments}
In this section, we present and analyze the experimental results.\footnote{Our code is available at https://github.com/zjukg/AnKGE} We first introduce the experimental settings in detail. Then we show the effectiveness and compatibility of the AnKGE with multiple base KGE models. Besides, we further analyze the effect of three levels analogical inference by ablation study. Finally, we conduct case study presenting a new view for the explanations of knowledge graph inference by analogical inference.

\subsection{Experiments Setup}

\subsubsection{Dataset}
We conduct experiments on link prediction task on two well-known benchmarks: WN18RR and FB15k-237. WN18RR and FB15k-237 are subsets of WN18 and FB15k, respectively. Some previous work \cite{DBLP:ConvE} has indicated the test leakage flaw in WN18 and FB15k, which means test triples appear in train dataset with inverse relations. WN18RR and FB15k-237 removing inverse relations are the modified version. Therefore, we use WN18RR and FB15k-237 as the experiment datasets. The statistic details of these datasets are summarized in Appendix \ref{app:Dataset}.

\subsubsection{Evaluation protocol}
We evaluate the KGE framework performance by four frequent evaluation metrics: the reciprocal mean of correct entity ranks in the whole entity set (MRR) and percentage of test triples with correct entities ranked in top 1/3/10 (Hit@1, Hit@3, Hit@10). For a test task $(h,r,?) \rightarrow t$, we replace all entities to create corrupted triples. Following the filter setting protocol, we exclude the other true triples appearing in train, valid and test datasets. Finally, we sort the filter corrupted triples according to the triple scores.

\subsubsection{Implementation details}
We train AnKGE framework based on four representative KGE models : TransE \cite{DBLP:TransE}, RotatE \cite{DBLP:RotatE}, HAKE \cite{DBLP:HAKE} and PairRE \cite{DBLP:PairRE}. We use the grid search to select the hyper-parameters of our framework. We search the number of analogy objects of three levels $N_e$, $N_r$ and $N_t$ $\in$ \{1, 3, 5, 10, 20\}, the basic weight of three levels $\alpha_E$, $\alpha_R$ and $\alpha_T$ $\in$ \{0.01, 0.05, 0.1, 0.2, 0.3\}, learn rate $\alpha \in \{1e^{-3}, 1e^{-4}, 1e^{-5}\}$. The loss function weight $\gamma$ in Equation (\ref{equ: level loss}) is set to 10, the transformation matrix weight $\lambda$ in Equation (\ref{equ: ent ana fun}) is set to 1 and 0 in FB15k-237 and WN18RR respectively. Before training AnKGE, we retrieve analogical objects of three levels in train dataset for once.
In both training and inference processes, AnKGE is extended based on the scoring function of the original model. Thus, AnKGE has the same model complexity as the original model. 

\subsection{Link Prediction Results}
\begin{figure}[t]
    \centering
\includegraphics[scale=0.5]{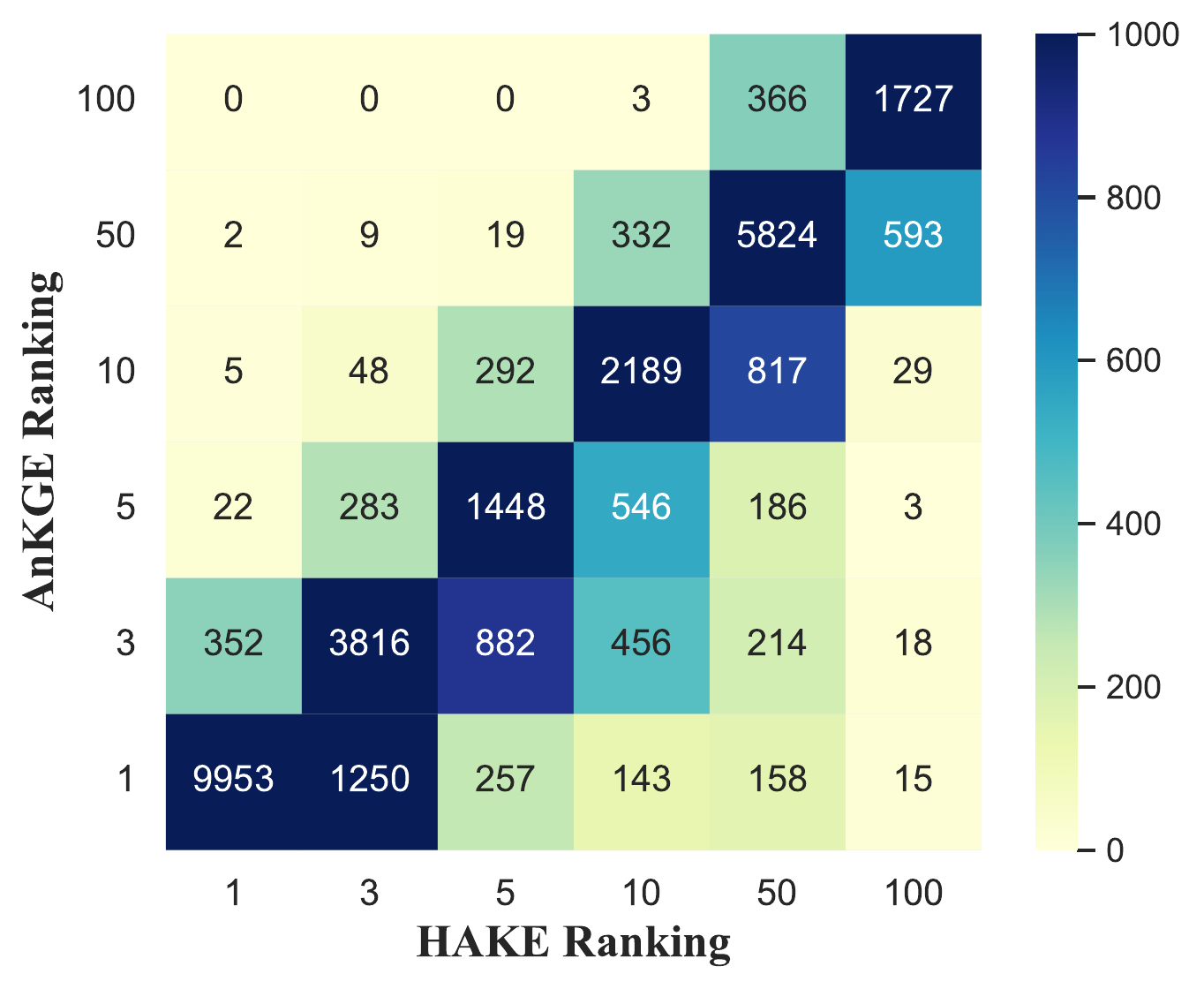}
	\caption{Comparison of the ranking between AnKGE and base model on the FB15k-237.}
	\label{fig:analyse}
\end{figure}

\begin{table}[t]
\centering

\begin{tabular}{lll|ll}
\toprule
\small
\multirow{2}{*}{\textbf{Models}} & \multicolumn{2}{c}{\textbf{FB15k-237}} & \multicolumn{2}{c}{\textbf{WN18RR}} \\ \cmidrule(lr){2-3} \cmidrule(lr){4-5}
& \multicolumn{1}{c}{MRR} & \multicolumn{1}{c}{Hit@1} & \multicolumn{1}{c}{MRR} & \multicolumn{1}{c}{Hit@1}  \\ \midrule
\textbf{AnKGE}
& \textbf{0.385} & \textbf{0.288} & \textbf{0.500} & 0.454 \\

w/o entity-level
& 0.384 & \textbf{0.288} & 0.497 & 0.451 \\

w/o relation-level
& 0.349 & 0.253 & \textbf{0.500} & \textbf{0.455} \\

w/o triple-level
& 0.384 & 0.287 & 0.499 & 0.453 \\

w/o all
& 0.349 & 0.252 & 0.496 & 0.452 \\

\bottomrule
\end{tabular}
\caption{Ablation study of three analogy level, where w/o means removing the corresponding level in AnKGE.}
\label{tab:ablation study}
\end{table}

\begin{table*}[t]
\centering

\resizebox{\textwidth}{!}{
\begin{tabular}{|c|c|c|c|c|}\hline

& \multirow{2}{*}{\textbf{Incomplete triple}} & \multirow{2}{*}{\textbf{Analogy object}} & \multirow{1}{*}{\textbf{AnKGE}} & \multirow{1}{*}{\textbf{Original}} \\
& & & \textbf{Rank} & \textbf{Rank} \\\cline{1-5}

\multirow{3}{*}{\textbf{Entity} }& (\textcolor[rgb]{0,0,0}{diencephalon}, \_has\_part, ?) $\rightarrow$ \emph{hypothalamus}  & brain & \textbf{5} & 25   \\ \cline{2-5}
\multirow{3}{*}{ }& (\textcolor[rgb]{0,0,0}{rest}, \_derivationally\_related\_form, ?) $\rightarrow$ \emph{breath}  & drowse & \textbf{6} & 38  \\ \cline{2-5}
\multirow{3}{*}{ }& (\textcolor[rgb]{0,0,0}{roof}, \_hypernym, ?) $\rightarrow$ \emph{protective\_covering}  & cap & 39 & \textbf{20}   \\ \hline

\multirow{3}{*}{\textbf{Relation}}& (felidae, \textcolor[rgb]{0,0,0}{\_member\_meronym}, ?) $\rightarrow$ \emph{panthera} & \_has\_part & \textbf{5} & 17  \\ \cline{2-5}
\multirow{3}{*}{ }& (monodontidae, \textcolor[rgb]{0,0,0}{\_member\_meronym}, ?) $\rightarrow$ \emph{delphinapterus} & \_hypernym\_Reverse & \textbf{1} & 64  \\ \cline{2-5}
\multirow{3}{*}{ }& (literary\_composition, \textcolor[rgb]{0,0,0}{\_hypernym}, ?) $\rightarrow$ \emph{writing} & \_has\_part & 88 & \textbf{18}  \\ \hline

\multirow{3}{*}{\textbf{Triple}}& (\textcolor[rgb]{0,0,0}{ticino, \_instance\_hypernym}, ?) $\rightarrow$ \emph{swiss\_canton} & (switzerland, \_has\_part) & \textbf{8} & 54  \\ \cline{2-5}
\multirow{3}{*}{ }& (\textcolor[rgb]{0,0,0}{south\_korea, \_has\_part}, ?) $\rightarrow$ \emph{inchon} & (port, \_instance\_hypernym\_Reverse) & \textbf{1} & 31  \\ \cline{2-5}
\multirow{3}{*}{ }& (\textcolor[rgb]{0,0,0}{elementary\_geometry, \_hypernym}, ?) $\rightarrow$ \emph{geometry} & (construct, \_synset\_domain\_topic\_of)  & 39 & \textbf{12}   \\ \hline

\end{tabular}
}

\caption{Analogical inference case Study. The better ranks are \textbf{blod}.}
\label{tab:case study}
\end{table*}

\subsubsection{Main results}
We use HAKE \cite{DBLP:HAKE} as the base model for AnKGE to compare with other baselines.
Baselines are selected from three categories \textbf{Conventional KGE models} including TransE \cite{DBLP:TransE}, ANALOGY \cite{DBLP:Analogy}, RotatE \cite{DBLP:RotatE}, HAKE,  Rot-Pro \cite{DBLP:Rot-Pro}, PairRE \cite{DBLP:PairRE}, and DualE \cite{DBLP:Dual}, \textbf{GNN-based KGE models} including R-GCN \cite{DBLP:R-GCN}, A2N \cite{DBLP:A2N},  CompGCN \cite{DBLP:CompGCN}, and SE-GNN \cite{DBLP:SE-GNN}, and \textbf{Enhanced KGE framework} including CAKE \cite{DBLP:CAKE}, PUDA \cite{DBLP:PUDA}, and REP \cite{DBLP:REP}.

The Table \ref{tab:link-prediction-sota} summarizes experiment results on FB15k-237 and WN18RR. The result of ANALOGY is from code\footnote{https://github.com/thunlp/OpenKE}.
The result of TransE, RotatE, HAKE and PairRE are from our trained model. The base model and AnKGE framework training details are provided in Appendix \ref{app:Training}. The other results are from the published paper. We can see that AnKGE enhances the analogical inference ability of the base model HAKE through analogical inference and outperforms the baseline models on most evaluation metrics except the Hit@10 metric where results of AnKGE slightly lower than PUDA and REP and achieve the second best. 
Overall, AnKGE remakes the rank of hard-predicted triples in HAKE by analogical inference, achieving the best results on both datasets.

\subsubsection{Compatibility results}
The AnKGE is a framework to enhance the analogical inference ability of KGE models, which retrieves analogical objects through $f_{kge}$ predefined in KGE models. Theoretically, our framework is applicable to most KGE models defining a score function for triples.
We chose four C-KGE models: TransE, RotatE, HAKE, PairRE as base model to validate compatibility.
As Table \ref{tab:ankge-mult-model} shows, AnKGE achieves a significant improvement over the base model on all metrics. The MRR metric improves by about 3\% on the FB15k-237. The result demonstrates that AnKGE is compatible with a wide range of KGE models. Moreover, AnKGE based on HAKE achieves a more significant improvement on FB15k-237 dataset. HAKE makes the entities hierarchical by using the depth of the entity to model different levels of the hierarchy, which is more helpful for analogical inference.

Compared with WN18RR, the improvement of the model on FB15k-237 is more significant, which we speculate is because FB15k-237 has richer relational patterns. So it has more improvement in the process of relation-level analogical inference. In addition, AnKGE is designed to predict the hard-predicted triples. The overall accuracy of FB15k-237 is lower than WN18RR. Consequently, the boosting effect of the model is reflected more obviously.

\subsection{Model Analysis}
\subsubsection{Ranking study}
In order to analyze the improvement effect of AnKGE, we compare the ranking results in FB15k-237 of the AnKGE-HAKE and original HAKE in Figure \ref{fig:analyse}. The horizontal coordinate represents the ranking range of the HAKE model, and the vertical coordinate represents the ranking range of AnKGE. We found that ranking changes are less apparent when the ranking is more significant than 100, so we selected the triples ranking within 100 and divided them into six ranking ranges for analysis. The diagonal line represents the unchanged ranking, the lower right of the diagonal line represents the AnKGE ranking as better than the HAKE ranking, and the upper left represents worse. We find some triples with worse rankings, but the number is much smaller than those with better rankings. In addition, the change in ranking is not so evident as the base model ranking increases; the better the base model ranking is, the more possible that AnKGE could improve the rankings.

\subsubsection{Ablation Study}
We conduct ablation experiments for the analogical inference part of AnkGE. Table \ref{tab:ablation study} shows the results of the ablation study for the AnKGE-HAKE on two datasets. We can see that the removal of any part makes the model less effective, except the relation-level on WN18RR dataset. Since there are only 11 relations in WN18RR, it is hard to retrieve suitable relation-level analogical objects. We explain this in more detail in case study.
In addition, the WN18RR consists of a lexicon containing contextual words that naturally provide entity-level analogical objects, which makes the model more effective for entity-level analogical inference. The result of FB15k-237 is the opposite. It may be because it has rich relationship patterns, making the relation-level analogical inference more effective.

\subsubsection{Case Study}
\label{sec: case study}
Analogical inference can generate explanations for predicted triples, which are valuable for real-life applications. Our method also provides an analogy view for the explanations of knowledge graph inference.
As the Table \ref{tab:case study} shows, we provide an intuitive demonstration about analogical inference. For each level, we select multiple example cases from WN18RR test set, and list their corresponding analogical objects and prediction results based on RotatE.
For entity-level, the idea is to retrieve hypernym or hyponym as the analogy object. For example, the diencephalon is located in the core of the brain. The fact that hypothalamus is part of brain improves the reliability of the people`s trust on predicted result. However, if hyponym entity becomes the analogy object, it will generate bad explanations and results. For instance, although cap can be regraded as a special type of roof, it is not the protective\_covering. Thus the misleading explanation that $(cap, \_hypernym, protective\_covering)$ downgrades the trustworthiness of the predicting result, which ranks the correct answer at 39.
For relation-level, AnKGE tends to retrieve the conceptually similar relations, such as the $(\_member\_meronym)$ and $(\_has\_part)$. Nevertheless, there are only 11 relations on WN18RR, which makes the AnKGE sometimes retrieve the inappropriate analogy relations. For example, $(\_hypernym)$ and $(\_has\_part)$ are the relations of opposite concepts, which leads to bad explanation and worse ranking. For triple-level, AnKGE typically focuses on the $(h,r)$ pair structure. As proof, ticino is a canton of Switzerland means that triple $(switzerland, \_has\_part, swiss\_canton)$ is good explanation.
However, sometimes the $(h,r)$ pair structure varies too much leading the misclassification. 

\section{Conclusion}
In this paper, we resort to analogical inference to study the knowledge graph completion task. We propose an analogical object retriever that retrieves appropriate analogical objects from entity-level, relation-level, and triple-level. Then, we design a novel and general self-supervised framework to enhance well-trained KGEs with analogical inference capability called AnKGE. Our method achieves competitive results on knowledge graph completion task and performs enhanced analogical inference ability. Some future directions include exploring more analogy patterns and a more general framework to adapt to the GNN-based KGE.

\section{Acknowledgments}
This work is funded by NSFCU19B2027/91846204.

\bibliography{aaai23}

\clearpage

\appendix

\begin{table}[!htpb]
    \centering
    \resizebox{1.0\columnwidth}{!}{
    \begin{tabular}{l *{5}{c}}
        \toprule
        Dataset & $|\mathcal{E}|$ & $|\mathcal{R}|$ & Train & Valid & Test \\
        \midrule
        FB15k-237 & 14,541 & 237 & 272,115 & 17,535 & 20,466 \\
        WN18RR & 40,493 & 11 & 86,835 & 3,034 & 3,134 \\
        \bottomrule
    \end{tabular}
    }
    \caption{Statistics of datasets. Train, Valid, and Test denote the size of train set, validation set, and test set, respectively.}
    \label{table: datasets}
\end{table}

\begin{table*}[ht]
    \centering
    \resizebox{2.1\columnwidth}{!}{
    \begin{tabular}{lccc}
        \toprule
        \textbf{Model  $\mathcal{M}$} & \textbf{ $f_{kge}(\textbf{h},\textbf{r},\textbf{t})$}& \textbf{$g_{kge}(\textbf{h},\textbf{r})$} & \textbf{Parameters}\\
        \midrule
        TransE \    & $-\|\textbf{h}+\textbf{r}-\textbf{t}\|_1$  & $\textbf{h}+\textbf{r}$  & $\textbf{h}, \textbf{r}, \textbf{t}\in\mathbb{R}^k$\\

        
        RotatE \     &$-\|\textbf{h}\circ \textbf{r}-\textbf{t}\|_2$ & $\textbf{h}\circ \textbf{r}$ & $\textbf{h}, \textbf{r}, \textbf{t}\in\mathbb{C}^k$, $|r_i|=1$\\
        
        \multirow{2}*{HAKE}    
        &$-\|\textbf{h}_m\circ \textbf{r}_m-\textbf{t}_m\|_2-$  &Cat$ \textbf{[}\|\textbf{h}_m\circ \textbf{r}_m\|_2,$ 
        &$\textbf{h}_m, \textbf{t}_m\in\mathbb{R}^k$,$\textbf{r}_m\in\mathbb{R}_+^k$, \\
        &$\lambda\|\sin((\textbf{h}_p+\textbf{r}_p-\textbf{t}_p)/2)\|_1$  &$\lambda\|\sin((\textbf{h}_p+\textbf{r}_p)/2)\|_1 \textbf{]}$
        &$\textbf{h}_p, \textbf{r}_p, \textbf{t}_p\in[0,2\pi)^k$, $\lambda\in\mathbb{R}^k$ \\
        
        PairRE \     &$-\|\textbf{h}\circ \textbf{r}^H-\textbf{t}\circ \textbf{r}^T\|_1$ & $ \textbf{h}\circ \textbf{r}^H $ & $\textbf{h}, \textbf{r}, \textbf{t}\in\mathbb{R}^k$\\
        
        \bottomrule
    \end{tabular} 
    }
    \caption{The details of knowledge graph embedding models, where $\| \cdot \|_1$ denotes the absolute-value norm, Cat ${\mathbf{[\cdot]}}$ denotes the concatenate vector function.}
    \label{table: model function}
\end{table*}

\begin{table*}[ht]
\centering

\resizebox{\textwidth}{!}{
\begin{tabular}{cccccccc}\hline

\multirow{2}{*}{\textbf{Dataset}} &\multirow{2}{*}{\textbf{Model}} & \multirow{1}{*}{\textbf{Emebdding}} & \multirow{2}{*}{\textbf{Margin}} & \multirow{1}{*}{\textbf{Adversarial}} & \multirow{1}{*}{\textbf{Negative}} & \multirow{2}{*}{\textbf{Batch Size}} & \multirow{2}{*}{\textbf{Inverse Relation}} \\
& & \textbf{Dimension} & & \textbf{Temperature} & \textbf{Samples} & & \\\cline{1-8}

\multirow{4}{*}{FB15k-237}& TransE & 500 & 9.0 & 1.0 & 256 & 1024 & True   \\ 
\multirow{4}{*}{         }& RotatE & 500 & 9.0 & 1.0 & 256 & 1024 & True   \\ 
\multirow{4}{*}{         }& HAKE & 1000 & 9.0 & 1.0 & 512 & 1024 & True   \\ 
\multirow{4}{*}{         }& PairRE & 1500 & 6.0 & 1.0 & 256 & 1024 & True   \\ \hline

\multirow{4}{*}{WN18RR   }& TransE & 500 & 6.0 & 1.0 & 256 & 2048 & True   \\ 
\multirow{4}{*}{         }& RotatE & 500 & 6.0 & 0.5 & 1024 & 512 & True   \\ 
\multirow{4}{*}{         }& HAKE & 500 & 6.0 & 0.5 & 1024 & 512 & True   \\
\multirow{4}{*}{         }& PairRE & 500 & 6.0 & 0.5 & 1024 & 512 & True   \\ \hline

\\ \hline

\multirow{2}{*}{\textbf{Dataset}} &\multirow{2}{*}{\textbf{Model}} & \multirow{1}{*}{\textbf{Entity}} & \multirow{1}{*}{\textbf{Relation}} & \multirow{1}{*}{\textbf{Triple}} & \multirow{1}{*}{\textbf{Entity}} & \multirow{1}{*}{\textbf{Relation}} & \multirow{1}{*}{\textbf{Triple}} \\
&  & \textbf{Cand.}$~N_e$ &\textbf{Cand.}$~N_r$ & \textbf{Cand.}$~N_t$ & \textbf{lambda}$~\alpha_E$ & \textbf{lambda}$~\alpha_R$  & \textbf{lambda}$~\alpha_T$ \\\cline{1-8}

\multirow{4}{*}{FB15k-237}& AnKGE-TransE & 1 & 1 & 3 & 0.01 & 0.2 & 0.02   \\ 
\multirow{4}{*}{         }& AnKGE-RotatE & 1 & 1 & 5 & 0.01 & 0.2 & 0.05   \\ 
\multirow{4}{*}{         }& AnKGE-HAKE & 1 & 1 & 5 & 0.05 & 0.3 & 0.1   \\ 
\multirow{4}{*}{         }& AnKGE-PairRE & 1 & 1 & 3 & 0.01 & 0.3 & 0.05   \\ \hline

\multirow{4}{*}{WN18RR   }& AnKGE-TransE & 1 & 1 & 20 & 0.01 & 0.3 & 0.3   \\ 
\multirow{4}{*}{         }& AnKGE-RotatE & 1 & 1 & 3 & 0.1 & 0.05 & 0.1   \\ 
\multirow{4}{*}{         }& AnKGE-HAKE & 1 & 1 & 3 & 0.1 & 0.05 & 0.1   \\
\multirow{4}{*}{         }& AnKGE-PairRE & 1 & 1 & 3 & 0.1 & 0.05 & 0.2   \\ \hline

\end{tabular}
}

\caption{The hyper-parameter settings of base model and AnKGE over different datasets.}
\label{tab:parameter}
\end{table*}

\section{KGE Models}
\label{app:score function}
We can divide knowledge graph embedding models into two categories: conventional knowledge graph embedding models and GNN-based models. Theoretically, the AnKGE framework applies to most conventional KGE models defining a score function for triples. In order to demonstrate the effectiveness and compatibility of AnKGE. We chose four representative conventional knowledge graph embedding models: TransE \cite{DBLP:TransE}, RotatE \cite{DBLP:RotatE}, HAKE \cite{DBLP:HAKE} and PairRE \cite{DBLP:PairRE} as base models for AnKGE. Table \ref{table: model function} exhibits the $g_{kge}(\cdot,\cdot)$ and $f_{kge}(\cdot, \cdot, \cdot)$ defined in these knowledge graph embedding models.
We will introduce the four KGE models, respectively.

\subsubsection{TransE} is the first knowledge graph embedding model proposing a geometric interpretation of the latent space. The TransE model is inspired by Word2vec vectors, requiring the sum of head embedding and relation embedding close to the tail embedding. It makes TransE successfully capture the relations between words through translations. However, due to the limit of translation, TransE cannot correctly handle N-to-one, one-to-N, and symmetric relations.

\subsubsection{RotatE} requires the embedding of $\mathbf{(h,r,t)}$ belong to $\mathbb{C}^k$, and considers the relation embedding as rotation vector in a complex latent space. Specifically, the complex component conveys the rotation along that axis, whereas the real component always equals 1. RotatE has been demonstrated that rotation allows modeling correctly numerous relational patterns, such as symmetry, anti-symmetry and inversion. However, RotatE cannot model the relation with hierarchy pattern.

\subsubsection{HAKE} is a hierarchy-aware knowledge graph embedding model that uses the depth of entity to model different levels of the hierarchy. HAKE distinguishes the entities into two categories: the entities at different levels of the hierarchy and the entities at the same level of the hierarchy, to model the semantic hierarchies. Experiments demonstrate that HAKE can effectively model the semantic hierarchies in knowledge graphs.

\subsubsection{PairRE} is a method capable of simultaneously encoding complex relations and multiple relation patterns. The model uses paired relation representations to adjust the margin in loss function to fit different complex relations. PairRE captures the semantic connection among relation vectors which have been demonstrated to encode three important relation patterns, symmetry/anti-symmetry, inversion and composition.

\section{Loss Weight}
\label{app:loss}

Considering the different contributions of the three levels, we introduce $\beta_E$, $\beta_R$ and $\beta_T$ to adjust gradient descent. Similarity to the softmax function, firstly, we replace the original element embedding with three level aggregated embeddings respectively, then calculate the exponential sum of analogy triple scores and original triple score, which is formulated as:
\begin{equation}
   \begin{split}
      \mathcal{T} = & e^{f_{kge}(\mathbf{h^+,r,t})}+e^{f_{kge}(\mathbf{h,r^+,t})} \\ & +e^{f_{kge}(\mathbf{z_e^+,z_r^+,t})}+e^{f_{kge}(\mathbf{h,r,t})}.
   \end{split}
\end{equation}
The weights of loss function in Equation (\ref{loss function}) are the rate of $\mathcal{T}$:
\begin{equation}
   \begin{split}
   &\beta_E = e^{f_{kge}(\mathbf{h^+,r,t})} / \mathcal{T}, \\
   &\beta_R = e^{f_{kge}(\mathbf{h,r^+,t})} / \mathcal{T}, \\
   &\beta_T = e^{f_{kge}(\mathbf{z_e^+, z_e^+,t})} / \mathcal{T}. \\
   \end{split}
\end{equation}
The highest analogy triple score means mapping original element embedding to aggregate embedding is necessary. If the original triple score is the highest, the triple should not analogy with other objects.

\section{Datasets}
\label{app:Dataset}
We evaluate the AnKGE framework on two widely-used datasets: WN18RR and FB15k-237. FB15k-237 is from the Freebase knowledge graph project, whose design is inspired by broadly used information communities such as The Semantic Web and Wikipedia. FB15k-237 contains information including locations,  media, geographical and people. WN18RR is from the WordNet knowledge graph project, a dataset that characterizes associations between English words.WN18RR contains information including symmetric, asymmetric and compositional relations. Statistics of these datasets are shown in Table \ref{table: datasets}.

\section{Implementation Details}
\label{app:Training}
Firstly, we train four KGE models: TransE, RotatE, HAKE and PairRE, as base models.
In the training stage, we apply a widely used negative sampling loss function with self-adversarial training:
\begin{align*}
    L=&\log\sigma(\gamma_m-f_{kge}(\textbf{h},\textbf{r},\textbf{t}))\\&+\sum_{i=1}^np(h'_i,r,t'_i)\log\sigma(f_{kge}(\textbf{h}'_i,\textbf{r},\textbf{t}'_i)-\gamma_m),
\end{align*}
where $\gamma_m$ is a fixed margin, $\sigma$ is the sigmoid function, $(h'_i,r,t'_i)$ is the $i$th corrupting negative triple for $(h,r,t)$ and $n$ is the number of negative triples. Moreover, $p(h'_j,r,t'_j)$ is the self-adversarial weight for this negative triple. The calculation of the weight is:
\begin{align*}
    p(h'_j,r,t'_j)=\frac{\exp \alpha_{temp} f_{kge}(\textbf{h}'_j,\textbf{r}, \textbf{t}'_j)}{\sum_i \exp \alpha_{temp} f_{kge}(\textbf{h}'_i,\textbf{r}, \textbf{t}'_i)}
\end{align*}
which is the probability distribution of negative sampling triples, where $\alpha_{temp}$ is the adversarial temperature of sampling.
When training and testing, we add reverse relations to expand the dataset. Specifically, for a triple $(h,r,t)$, we add a new reverse triple $(t, r^{-1}, h)$ in dataset. $r^{-1}$ represents the reverse relation of $r$. In the link prediction task, the model only predicts tail entity, which is equivalent to the effect of predicting both head and tail entities.

Then, we use AnKGE to enhance the well-trained KGEs with analogical inference capability. We show that AnKGE achieves competitive results on knowledge graph completion task and performs enhanced analogical inference ability. The loss function weight $\gamma$ in Equation (\ref{equ: level loss}) is set to 10, the transformation matrix weight $\lambda$ in Equation (\ref{equ: ent ana fun}) is set to 1 and 0 in FB15k-237 and WN18RR respectively. We use the grid search to select other hyper-parameters, including: entity candidates $N_e$, relation candidates $N_r$ triple candidates $N_t$, entity lambda $\alpha_E$, relation lambda $\alpha_R$ and triple lambda $\alpha_T$. Other experimental settings are the same. The experiment of AnKGE-TransE on WN18RR is the only exception. We use fixed weight parameter instead of adaptive weight parameter and cosine similarity instead of euclidean norm.

In addition, since there is no negative sampling, the memory footprint and time cost are lower than the base model, which is generally acceptable.

We implement all the models with PyTorch, and run experiments on NVIDIA RTX3090 GPUs with 24GB RAM and Intel(R) Xeon(R) Silver 4210R CPU @ 2.40GHz with 40 cores. The hyper-parameter settings of base model and AnKGE are shown in Table \ref{tab:parameter}

\end{document}